\documentclass[lettersize,journal]{IEEEtran}
\usepackage{cite}
\usepackage{amsmath,amssymb,amsfonts}
\usepackage{algorithmic}
\usepackage{graphicx,color}
\usepackage{textcomp}
\def\BibTeX{{\rm B\kern-.05em{\sc i\kern-.025em b}\kern-.08em
    T\kern-.1667em\lower.7ex\hbox{E}\kern-.125emX}}
\AtBeginDocument{\definecolor{ojcolor}{cmyk}{0.93,0.59,0.15,0.02}}

\usepackage{array}
\usepackage{textcomp}
\usepackage{stfloats}
\usepackage{url}
\usepackage[table]{xcolor}
\definecolor{stoprow}{RGB}{253,235,208}
\usepackage{cite}
\begin{document}
\title{A Closed-Loop Evaluation of Capability Loss and Recovery in Compressed Driving Policies}

\author{Ahmad Alfan Alfian Irfan$^{1}$, Nur Ahmad Khatim$^{2}$, Mansur Arief$^{3}$

\thanks{Ahmad Alfan Alfian Irfan is with the Information Technology, Universitas Muhammadiyah Yogyakarta, Indonesia}
\thanks{Nur Ahmad Khatim is with the Division of Information Science, Nara Institute of Science and Technology (NAIST), Japan}
\thanks{Mansur Arief is with the Industrial and Systems Engineering (ISE) Department and IRC for Smart Mobility and Logistics (IRC-SML), King Fahd University of Petroleum and Minerals (KFUPM), Saudi Arabia}
\thanks{This work was supported by the IRC for Smart Mobility and Logistics (IRC-SML), King Fahd University of Petroleum and Minerals (KFUPM), Saudi Arabia through research grant INML2654. }
}

\maketitle

\begin{abstract}
Many automobile and mobility companies deploy learned driving policies on
embedded computers with limited memory and power. Pruning, knowledge
distillation, and quantization are the standard methods to reduce the size and
the inference cost of these policies. However, these methods are commonly
assessed by aggregate numerical scores, and such scores may not reflect the
ability of the policy to drive safely when interacting with other road users. In this study, we propose a stage-wise closed-loop evaluation approach to follow a driving policy through a compression
pipeline. We formulate the driving task as a partially observable Markov
decision process (POMDP) and train a belief-state policy with proximal policy
optimization (PPO) in Gym-Duckietown. We then extract the actor, compress it
one stage at a time, and evaluate it on five driving curricula. We show that structured pruning is the
stage at which the driving capability is first lost. Meanwhile, distillation improves the
pruned actor, but the improvement is limited by its rehearsal data. 
Integer quantization of the improved actor loses some of the curricula that require
the vehicle to stop and then resume. Interestingly, the same procedure on the unpruned actor
preserves all five curricula. Our study thus provides an empirical analysis aiming to 
answer the currently active discussions on how to accept a compressed driving
policy, so as to achieve a safe and statistically reliable deployment of
automated driving functions.
\end{abstract}


\section{INTRODUCTION}

\IEEEPARstart{M}{any} of the industrial leaders in automobiles and mobility
providers now use learned policies for perception and decision-making in their
vehicles. These policies must run on embedded computers, whose memory, power,
and thermal limits are set by vehicle cost. Thus, a trained network is commonly
compressed before it is deployed, where pruning removes network capacity
\cite{blalock2020pruning,li2017pruningfilters}, quantization reduces numerical
precision \cite{jacob2018quantization,nagel2021whitepaper}, and knowledge
distillation is applied later to recover what compression has lost
\cite{gou2021kdsurvey}. Progress in all three is normally measured by model
size, inference latency, and test-set accuracy.

However, aggregate accuracy can conceal what compression removes, since models
with very different parameter counts can agree on top-line metrics while
differing a lot on a small subset of inputs \cite{hooker2019forget}. For a
driving policy the problem is more critical. The main reason is that a policy
acts on its environment repeatedly, so that a small change in one action
changes the next observation, which can carry the vehicle onto a different
trajectory.
Thus, a compressed policy may stay close to its original action mapping and
still lose behaviors that were previously reliable, such as following the lane,
stopping at a line, resuming later, or yielding to a crossing pedestrian.
Compressing a driving policy is therefore a question of two parts, i.e. how far
the network can be reduced, and which learned capabilities survive each stage
of the optimization.

The interaction between these stages makes the problem harder, since pruning
can remove useful capacity, distillation can bring back part of the lost
behavior, and quantization may alter the improved policy again. It has been
shown for
supervised models that the order of these stages measurably changes the final
accuracy \cite{shen2024order,kim2026prunequantizeorder}. What those studies
optimize, however, is an accuracy score on a static test set. A policy that must
act raises four connected questions instead: (i) at which stage does the driving
capability first break, (ii) can it be recovered after pruning, (iii) does the
rehearsal data decide what is recovered, and (iv) does that recovery survive a
later reduction in precision.

Addressing the above challenges, our contribution is predominantly to develop a
stage-wise closed-loop evaluation approach for a driving policy under
compression. We study this approach on a visuomotor driving policy in
Gym-Duckietown \cite{paull2017duckietown,chevalierboisvert2018gymduckietown}. We
formulate the driving task as a Partially Observable Markov Decision Process (POMDP) and learn the policy with belief-state Proximal Policy
Optimization (PPO)
\cite{schulman2017ppo,ni2022recurrent}. We then isolate the actor network and
follow its behavior through a sequence of compression stages. Every stage is
evaluated in the driving curricula
against the same acceptance criteria.

As described more in later sections, our experiments reveal a clear loss-and-recovery pattern. Structured pruning
reduces the machine learning-based actor size to less than 10\% its original size, and the resulting
64-unit hidden layers lose the driving capability on all five curricula.
Knowledge distillation improves the pruned actor, although how much it improves
is decided by its rehearsal data, i.e. by the driving states on which the
student imitates the teacher. While distilling on a limited task distribution
recovers only part of the lost behavior, balanced rehearsal across the
curricula recovers the complete tested behavior of the floating point (FP) 32 actor. Further
compression changes the outcome again, since quantizing the pruned actor
directly fails every curriculum, and post-training quantization brings back the
task-level failure even after a successful distillation. These results suggest that a compression pipeline for a learned driving policy
should be studied as a sequence of behavioral transitions. A compressed actor
can lose its capability, regain it through appropriate rehearsal, and lose part
of it again at a later stage. Besides, we observe that action errors can stay
well inside our tolerance for numerical agreement while the driving capability
degrades. For this reason we judge a compressed policy by how it drives and not
by how closely it reproduces recorded actions.

Our contributions are threefold. First, we characterize the capability loss across a sequential compression
pipeline for a belief-based visuomotor driving policy. We trace the actor from
structured pruning through distillation to reduced-precision deployment. We also
formulate the evaluation as a scenario-based acceptance problem by varying the
system under test. Second, we show that the recovery after pruning depends on the rehearsal coverage.
Limited-task distillation recovers only part of the lost capability. Balanced
rehearsal can recover the complete tested behavior of the compressed FP32 actor. Finally, we demonstrate that a successful recovery does not guarantee robustness to
a later precision reduction. The tested INT8 routes bring back the task-level
failures, while FP16 preserves the recovered behavior. Thus, action-level
similarity alone is not sufficient to guarantee a preserved driving capability.

The rest of this work will be presented as follows. In Section~\ref{sec:related}
we position our study with respect to the compression literature and to the
scenario-based evaluation of automated driving. In Section~\ref{sec:setting} we
formulate the driving task and describe the belief-state policy model. In
Section~\ref{sec:pipeline} we specify the compression pipeline and the
evaluation protocol. In Section~\ref{sec:results} we present our numerical
experiment and findings. In Section~\ref{sec:discussion} we discuss the value of
the stage-wise view, its implications for deployment practice, and our limitations. We conclude in
Section~\ref{sec:conclusion}.

\section{RELATED WORK}
\label{sec:related}

In this section, we review the three compression methods we use, the evidence
that their stages interact, the application of compression to driving systems,
and the scenario-based evaluation literature.

\subsection{Pruning and Structured Sparsity}

Pruning reduces a network by removing the parameters that contribute least to
its output. The field is large enough that its benchmarking practices have
themselves been surveyed and criticized \cite{blalock2020pruning}. While
unstructured pruning zeroes individual weights and can reach high sparsity, the
irregular pattern it leaves does not by itself reduce computation
\cite{li2017pruningfilters}. Structured pruning instead removes whole filters
or units, so that the compressed network stays dense and runs faster on
ordinary hardware \cite{li2017pruningfilters,he2024structuredsurvey}. Criteria
for
deciding what to remove range from weight magnitude to Taylor-expansion
estimates of the contribution of a unit to the loss \cite{molchanov2019importance}.
This literature is developed largely on supervised classification, where the
effect of pruning is evaluated on the test-set accuracy. When the same methods are
carried into deep reinforcement learning, they have been reported to lose
considerable performance \cite{livne2020pops}. This is one reason we aim to pinpoint the
capability loss empirically in this study.

\subsection{Knowledge Distillation}

Knowledge distillation (KD) trains a compact student to reproduce the outputs of a
larger teacher \cite{gou2021kdsurvey, li2023taskkd}. It is the usual way to close the gap that
compression opens. For control policies the teacher supplies actions instead of
class scores. It has been shown that policy distillation can compress an agent
substantially while retaining expert-level play, and can fold several expert
policies into one multi-task student \cite{rusu2016policydistillation}. That
result establishes that the capability of a student depends on which experts and
which states it rehearses on. In our case we isolate that dependence within a
single task family. With the teacher, the loss, the optimizer, and the training
budget all held fixed, the same distillation procedure recovers either part or
all of the lost driving behavior, and the only difference is which driving
curricula its rehearsal states are drawn from.

\subsection{Quantization and Reduced Precision}

Quantization lowers the numeric precision of weights and activations. Integer
schemes are designed so that inference can run in integer arithmetic alone
\cite{jacob2018quantization}. Post-training quantization calibrates an already trained network, and it is usually sufficient to reach 8-bit accuracy close to floating point. Quantization-aware training instead simulates the reduced precision during optimization, and is normally reserved for lower bit widths \cite{nagel2021whitepaper}. Reduced floating-point formats are a separate route.
FP16 storage with wider accumulation is standard practice, and the need for FP32
accumulation in reductions and dot products is documented
\cite{micikevicius2018mixed,vanbaalen2023fp8}. Quantization has also been studied for reinforcement learning, where policies were reported to tolerate six to eight bits without loss of reward, and where quantization-aware training consistently outperformed post-training quantization \cite{krishnan2019quarl}. While our results agree with the first observation for an actor that has not been pruned, they diverge from both once the actor has been pruned and improved by distillation.
To the best of our knowledge, the literature is missing a comparison in which an integer route and a reduced
floating-point route start from the same fixed checkpoint. We study that
comparison here and evaluate them against identical task-level criteria in the experiments.

\subsection{Compression Stages for Driving Systems}

Compression stages are known not to be independent. For convolutional networks,
a systematic ordering over distillation, pruning, quantization, and early exit
has been derived, in which distillation is placed first and quantization near
the end \cite{shen2024order}. More recent work argues on theoretical grounds
that weaker perturbations should precede stronger ones, and concludes that
pruning before quantizing outperforms the reverse
\cite{kim2026prunequantizeorder}. Both studies are conducted on vision and
language models, where the quantity being ordered for is accuracy on a static
test set. Our work follows  the latter, since we prune before
quantizing. Distillation is done after pruning, i.e.
to improve a capability that has already been lost. We observe that the value of
distillation in this position is decided by the data it rehearses on. 

Compression has been applied to driving systems in order to meet onboard
compute limits, and some pipelines prune first and then use distillation to
recover the loss \cite{wang2025compressingmultitask}. Such work typically
targets perception models, and reports perception metrics, such as detection
and segmentation scores, together with frame rate. Sensor and compute limits
have likewise been treated as design variables, for instance in the optimal
placement of onboard LiDARs under an explicit cost constraint
\cite{liu2019lidar}. In the compression case, the compressed model is evaluated
only in its final form, which is suitable for a perception module whose outputs
do not feed back into its own future inputs. A driving policy, in contrast,
closes that loop. Thus, a score on
the final model cannot show which stage removed a behavior, which stage brought
it back, or whether a later stage reversed an earlier improvement.

\subsection{Scenario-Based Evaluation of Automated Driving}
\label{sec:related-scenario}

The question of how much testing is enough for an automated vehicle has been
studied extensively. Kalra and Paddock estimated the driving distance required
to demonstrate reliability at the population level, and concluded that the
requirement is simply unaffordable by direct on-road accumulation
\cite{kalra2016driving}. Scenario-based assessment has since become the
dominant response, and its variants have been surveyed and categorized in
several complementary ways \cite{riedmaier2020survey,ding2023survey, zhang2024accelerated}. A
parallel line of work accelerates the evaluation itself, either by skewing the
sampling distribution toward safety-critical events and correcting the estimate
later \cite{zhao2017accelerated,arief2021deep}, or by selecting the deployment
environment adaptively, so that fewer and safer trials suffice
\cite{arief2018accelerated}. The uncertainty of such data-driven testing has
also been quantified explicitly \cite{huang2019evaluation}. Besides, generative
models have been used to reconstruct the scenarios on which the
assessment is carried out \cite{guo2019gaussian}.

\section{PROBLEM FORMULATION}
\label{sec:setting}

In this section, we describe the driving task, its formulation as a POMDP, the
belief representation, and the actor that is compressed in the later stages.

\subsection{Driving Task and Evaluation Curricula}

\begin{figure*}[!t]
\centering
\includegraphics[width=\textwidth]{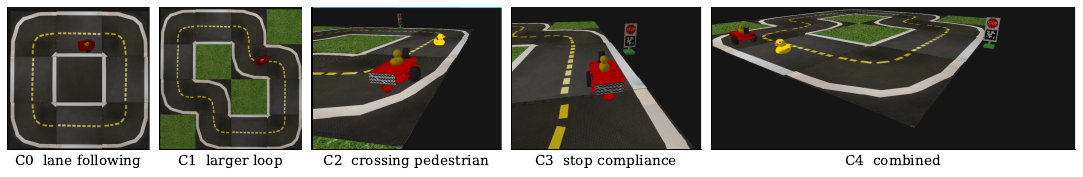}
\caption{The five driving curricula rendering through onboard camera frames taken
from evaluation rollouts in Gym-Duckietown.}
\label{fig:task}
\end{figure*}

The task is visuomotor lane-following with pedestrian avoidance and stop-line
compliance in Gym-Duckietown
\cite{paull2017duckietown,chevalierboisvert2018gymduckietown}. Capability is
organized as five driving curricula of increasing difficulty summarized in Table
\ref{tab:curricula} with visual rendering in Fig.~\ref{fig:task}. These curricula serve two roles at once, since during
training they are a curriculum in the ordinary sense, and during evaluation
they are the fixed scenario set described in
Section~II-\ref{sec:related-scenario}. C0 is plain lane following on a small loop
with domain randomization, C1 moves to a larger loop, and C2 adds crossing
pedestrians. C3 removes pedestrians but adds a stop sign, which requires the
policy to stop, hold, and then resume. C4 combines crossing pedestrians with the
stop sign on the same loop and doubles the episode horizon. The two later
curricula are the demanding ones, since they require the policy to interrupt
its own driving and then establish it again, and it is exactly these two that
reappear throughout our results.

\begin{table}[!t]
\caption{The driving curricula for both training and evaluation.}
\label{tab:curricula}
\centering
\small
\begin{tabular}{lllccr}
\hline
\textbf{ID} & \textbf{Map} & \textbf{Pedestrians} & \textbf{Stop} & \textbf{Random} & \textbf{Horizon} \\
\hline
C0 & small loop & no & no & yes & 1{,}900 \\
C1 & large loop & no & no & yes & 2{,}700 \\
C2 & large loop & crossing & no & no & 2{,}700 \\
C3 & large loop & no & yes & no & 2{,}700 \\
C4 & large loop & crossing & yes & no & 4{,}200 \\
\hline
\end{tabular}
\end{table}

\subsection{POMDP Formulation}

Each curriculum defines a POMDP
$(\mathcal{S}, \mathcal{A}, T, R, \Omega, O, \gamma)$. The state $s_t \in
\mathcal{S}$ is the simulator state, comprising the vehicle pose and velocity,
the lane geometry, and, where the curriculum activates them, the pedestrian and
stop-sign configurations. The action
\begin{equation}
    a_t = (u_v, u_\omega) \in \mathcal{A} = [-1,1]^2
\end{equation} 
maps linearly to a linear
velocity command of at most $0.4$\,m/s and a yaw-rate command spanning
$8$\,rad/s. The transition kernel $T(s_{t+1} \mid s_t, a_t)$ is the
differential-drive simulator dynamics together with pedestrian motion.
The observation $o_t \in \Omega$ is a monocular camera frame, from which a
policy must infer the state $s_t$ through its belief. Partial observability is
well motivated here, since pedestrians leave the field of view mid-crossing,
stop signs are visible long before the stop line and invisible at it, and the
detector both misses and false-fires. The setting remains simplified, so that
we can focus on how the belief-state representation is compressed and actuated
by the learned policy.

The reward is a fixed weighted sum of six terms, combining forward progress,
lane keeping, pedestrian proximity, stop compliance, action smoothness, and a
terminal bonus or penalty, with weights set per curriculum. The objective is the usual
discounted return. We emphasize that the
reward is used only to train the original policy, since our acceptance criteria
are defined on task outcomes directly.

\subsection{Perception and EKF-Based Belief}

Instead of conditioning a recurrent policy on raw frames, the system factors the
belief into interpretable components. The transformation from pixels to the
policy input has three steps, where metric measurements are first extracted
from the frame, the measurements are then fused over time by extended Kalman
filters (EKFs), and the filter posteriors are finally assembled into a fixed
29-dimensional vector serving as the policy input.

\subsubsection{From bounding boxes to metric measurements}
A fine-tuned YOLO11n detector \cite{ultralytics2024yolo} returns bounding boxes
with a class and a confidence score per frame. 
Detections first pass per-class confidence thresholds, which are 0.40 for
pedestrians and 0.10 for stop signs, both fixed by our preliminary calibration
study. The bottom-center pixel of the surviving box is cast as a ray through
the calibrated camera model and intersected with the ground plane, giving a
range 
\begin{equation}
    r = \sqrt{x^2 + y^2}
    \end{equation}
and a bearing 
\begin{equation}
    \theta = \operatorname{atan2}(x, y)
\end{equation} in the robot frame. A linear range
correction $r \mapsto ar + b$, fitted offline against calibration recordings,
removes the systematic projection bias, and the resulting polar measurement
carries a noise model that grows with range.

\subsubsection{From pixels to lane measurements}
A MobileNetV3-small regressor \cite{howard2019mobilenetv3} takes the resized
frame and outputs three numbers in normalized units, i.e. the lateral offset
$d$ in meters, the heading error $\varphi$ in radians, and the curvature
$\kappa$ in 1/m. Its measurement noise is taken to be the residual spread of
the regressor, measured offline on separate validation data.

\subsubsection{Belief initialization, prediction, and correction}
At episode reset the belief starts uncommitted, in the sense that the lane
filter is uninitialized with validity zero, the existence probability of each
object is set at its prior, and the stop mode is \emph{none}. The lane EKF
initializes on the first lane measurement, taking that measurement as its state
$[d, \varphi, \kappa]$ with covariance set at preset initial values. Later it alternates
kinematic prediction with the ego-motion estimate and standard EKF correction
by each new measurement, with process noise growing in $\Delta t$.

Each object runs its own EKF over planar position and velocity in the robot
frame. Then, prediction re-expresses the state in the new robot frame using the
ego-motion estimate, so that the tracked velocity stays consistent with the
physical motion of the pedestrian. The track initializes on the first accepted
detection with zero velocity and a wide velocity uncertainty, and corrections
are applied directly in polar coordinates. Existence is tracked separately by a Bernoulli filter with survival and birth
priors and a Bayes update using the calibrated detection and false-positive
rates of the detector. The Bayes step is skipped whenever the predicted
position lies outside the view of the camera. The ego-motion
estimate itself is obtained by differencing consecutive poses. We note that no lane geometry,
object position, or other privileged simulator state enters the belief.

\subsubsection{From posteriors to the 29 policy inputs}
The stop-obligation state machine, driven by the stop-sign belief and a route
prior, tracks whether a stop is currently not required, required, or already
satisfied, and supplies the distance to the stop line. The filter posteriors
are
then read out into the fixed 29-dimensional vector, shown in Table~\ref{tab:belief}. The belief here is a factored parametric approximation, where each factor is represented by its sufficient statistics, so the policy conditions on a finite-dimensional vector without loss of the modeled distributional information. 
The policy $\pi(a_t \mid b_t)$ is then trained on this vector with PPO
\cite{schulman2017ppo}. We follow that model-free training over a
sufficiently informative belief input is a strong baseline for POMDPs
\cite{ni2022recurrent}.

\begin{table*}[!t]
\caption{The components of the belief-state vector $b_t$. }
\label{tab:belief}
\centering
\small
\setlength{\tabcolsep}{5pt}
\begin{tabular}{cllll}
\hline
\textbf{\#} & \textbf{Component} & \textbf{Unit} & \textbf{Group} & \textbf{Produced By} \\
\hline
1 & lane validity probability & --- & lane belief & lane EKF, fed by MobileNetV3-small \\
2--3 & lateral error, mean and std & m & & \\
4--5 & heading error, mean and std & rad & & \\
6--7 & curvature, mean and std & 1/m & & \\
\hline
8 & actual linear velocity & m/s & egomotion & pose differencing (odometry proxy) \\
9 & actual yaw rate & rad/s & & \\
\hline
10 & stop-line distance & m & stop geometry & stop-sign belief and route prior \\
\hline
11 & pedestrian existence probability & --- & pedestrian belief & pedestrian EKF + existence filter, fed by YOLO11n \\
12--13 & range, mean and std & m & & \\
14--15 & bearing, mean and std & rad & & \\
16--17 & radial velocity, mean and std & m/s & & \\
18--19 & bearing rate, mean and std & rad/s & & \\
\hline
20 & stop-sign existence probability & --- & stop-sign belief & stop-sign belief updater, fed by YOLO11n \\
21--22 & range, mean and std & m & & \\
23--24 & bearing, mean and std & rad & & \\
\hline
25--27 & stop mode, one-hot: none / required / satisfied & --- & obligation & stop state machine \\
\hline
28 & previous linear velocity command & m/s & action feedback & policy output at $t{-}1$ \\
29 & previous angular velocity command & rad/s & & \\
\hline
\end{tabular}
\end{table*}

\subsection{Actor Extraction}

The trained actor is a three-layer perceptron with tanh activations, mapping
$b_t \in \mathbb{R}^{29}$ through two hidden layers of 256 units to the
2-dimensional action, for 73{,}986 parameters. Evaluation uses the
deterministic mean action, and everything downstream of this point compresses
the actor alone. The perception and belief stages are never modified, which
keeps the comparison interpretable but also bounds the system-level payoff,
since perception dominates the end-to-end cost. The question throughout is thus
whether the mapping from belief to action survives being made smaller.

\section{COMPRESSION AND EVALUATION PIPELINES}
\label{sec:pipeline}

\begin{figure*}[!t]
\centering
\includegraphics[width=\textwidth]{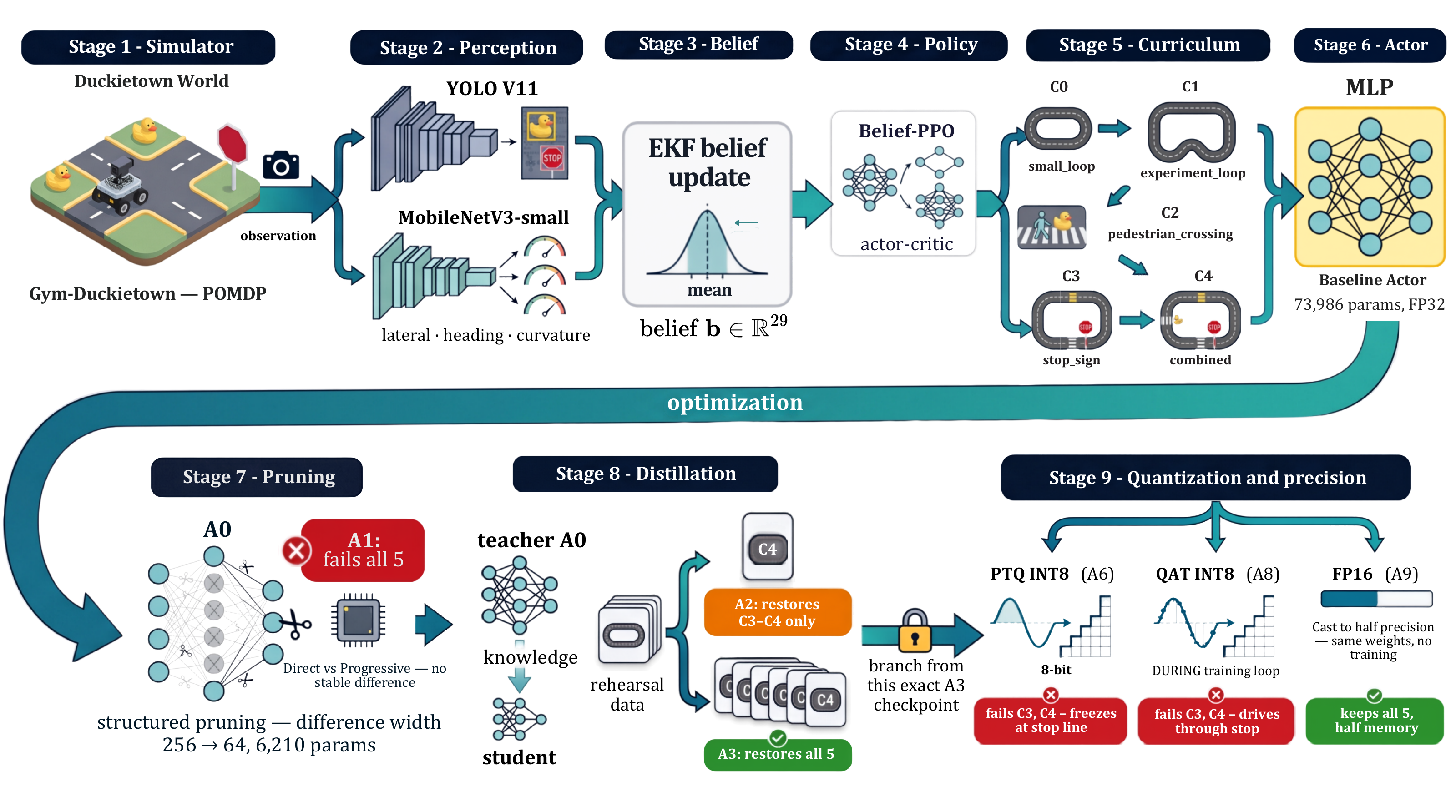}
\caption{Overview of the study. Top: the driving policy is built in
Gym-Duckietown, where YOLO and MobileNetV3-small measurements are fused by
EKF-based belief updaters into a belief state, and a Belief-PPO
actor-critic is trained through the five curricula. Bottom: the extracted actor moves through structured pruning, two
distillation branches that differ only in rehearsal data, and three precision
routes from the same checkpoint.}
\label{fig:pipeline}
\end{figure*}
In this section, we describe our compression and evaluation details.

\subsection{Pipeline Configurations}

Fig.~\ref{fig:pipeline} gives an overview of the whole study.
Table~\ref{tab:candidates} lists the ten evaluated models. A0 is the original
actor and serves as the reference on every curriculum. A1 through A8 realize
every meaningful placement of pruning, distillation, and quantization that the
framework supports. Two of them are controls, where A4 applies quantization
without pruning, and A5 applies pruning and quantization without distillation.
A9 is the reduced-floating-point control. A6
and A8 both descend from the byte-identical A3 checkpoint, and the difference
between them is the quantization route alone.

\begin{table}[!t]
\caption{Evaluated pipeline candidates.}
\label{tab:candidates}
\centering
\small
\setlength{\tabcolsep}{3pt}
\begin{tabular}{llcc}
\hline
\textbf{ID} & \textbf{Construction} & \textbf{Width} & \textbf{Precision} \\
\hline
A0 & original actor (reference) & 256 & FP32 \\
A1 & prune & 64 & FP32 \\
A2 & prune $\to$ KD(C4) & 64 & FP32 \\
A3 & prune $\to$ KD(bal) & 64 & FP32 \\
A4 & PTQ only, no pruning (control) & 256 & INT8 \\
A5 & prune $\to$ PTQ, no KD (control) & 64 & INT8 \\
A6 & prune $\to$ KD(bal) $\to$ PTQ & 64 & INT8 \\
A7 & prune $\to$ KD(C4) $\to$ PTQ $\to$ QAT(C4) & 64 & INT8 \\
A8 & prune $\to$ KD(bal) $\to$ QAT(bal) & 64 & INT8 \\
A9 & prune $\to$ KD(bal) $\to$ FP16 cast & 64 & FP16 \\
\hline
\end{tabular}
\end{table}

\subsection{Compression Stages}

The compression stages consist of pruning, distillation, and quantization (INT and FP16). To that end, structured pruning removes whole hidden units. Each unit is scored by the sum
of its L2 incoming and outgoing connectivity plus its absolute bias, ties keep
the lower index. Thus, the width-64
actor retains 6{,}210 of 73{,}986 parameters, yielding a 91.6\,\% reduction.

In terms of distillation, the original actor is kept fixed and serves as the teacher, while the student
minimizes a Smooth-L1 loss between the deterministic actions of the teacher and
of the student, normalized by the physical action ranges. We use phase-balanced
sampling, so that nominal driving cannot dominate the batch, and no
ground-truth simulator state enters training. The two rehearsal sets differ
only in coverage, where the historical set is drawn from C4 development states,
and the balanced set holds 62{,}176 public states drawn across all five
curricula. Teacher, loss,
optimizer, batch size, learning rate, and epoch budget are identical between the
two.

Our post-training quantization approach uses eager static INT8 on the x86 backend, with per-channel symmetric
weights and per-tensor affine activations, calibrated on development states
only \cite{jacob2018quantization,nagel2021whitepaper}.The A9 candidate casts the A3 weights to 16-bit floats and changes nothing
else. Since a CPU backend can silently widen half-precision arithmetic back to
FP32, A9 is admitted only after an explicit validity check, which confirms that
the forward pass executes natively in half precision and that the outputs are
FP16 and differ from the FP32 source
\cite{micikevicius2018mixed}.

\subsection{Closed-Loop Acceptance Criterion}

Every candidate drives every curriculum on the same eight seeds, giving 40
episodes per candidate and 400 in total, and acceptance is decided by checks
that were fixed before any result existed. Let $\mathcal{K}$ denote that set of
checks. A candidate $m$ passes curriculum $c$ if and only if
\begin{align}
\mathrm{PASS}(m,c) = \bigwedge_{k \in \mathcal{K}} & \Big[ q_k(m,c) \preceq \tau_k \nonumber \\
&\wedge \Delta_k\big(q_k(m,c), q_k(A0,c)\big) \preceq \delta_k \Big],
\label{eq:accept}
\end{align}
where $q_k$ is the measured quantity for check $k$ and $\tau_k$ is its absolute
requirement. Here $\delta_k$ is the largest regression tolerated relative to the
reference actor A0 on the same seeds, and $\preceq$ denotes the direction in
which each check is favorable. The checks cover completion, progress,
collisions, unsafe proximity, stop violations, stop completion and restart, lane
failures, invalid poses, and minimum pedestrian clearance.

We note two aspects of \eqref{eq:accept}. First, a curriculum
decision is PASS only if every check holds, so that a high completion count does
not by itself imply acceptance. Second, the paired term $\Delta_k(\cdot,
q_k(A0,c))$ means that a documented weakness of the original policy is not
charged to compression unless it worsens beyond the preset margin $\delta_k$.
We note that no threshold was changed after the
results were opened.

\section{NUMERICAL EXPERIMENTS AND RESULTS}
\label{sec:results}

\begin{figure}[!t]
\centering
\includegraphics[width=\linewidth]{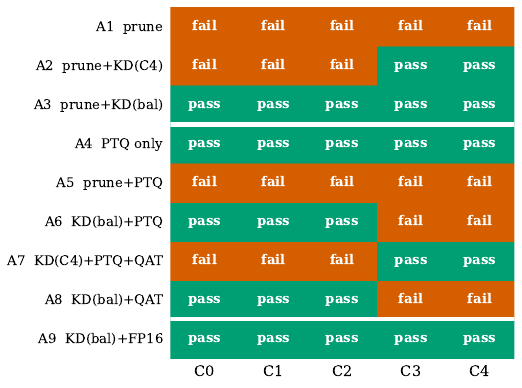}
\caption{Cross-curriculum decisions across the compression pipeline.}
\label{fig:matrix}
\end{figure}

In this section, we present the settings and findings from our numerical
experiments. Figure~\ref{fig:matrix} reports the decision per curriculum.

\subsection{Capability Loss under Pruning}

The pruned actor A1 fails every curriculum. On C0 through C2 the dominant
modes are invalid poses and lane failures, with completion falling to three,
three, and zero episodes of eight. On C3 every episode ends in a stop
violation. It is thus clear that the loss is neither partial nor specific to one
curriculum. Structured
pruning to width 64 is the first stage in this pipeline at which the task-level
capability is lost.
Table~\ref{tab:failures} records the exact state and action at the first failure of each failing candidate.

\subsection{Effect of the Rehearsal Coverage}

While distillation does improve the pruned actor, the extent of the improvement
is determined by the student rehearsal data. A2 and A3 share the
pruned source configuration, the teacher, the loss, and the training budget,
and differ only in rehearsal coverage. A2, rehearsing on C4-focused states,
completes every C3 and C4 episode and none of C0 through C2, where it shows
total invalid-pose or lane-failure rates. A3, rehearsing on balanced C0-C4
states, passes all five curricula and also passes every action-fidelity check.
In other words, expanding the coverage flipped three curricula from complete
failure to full recovery. 

\subsection{Placement of Distillation and Quantization}
\label{sec:quant-after-recovery}

A5 quantizes the pruned actor directly. It fails all five curricula, with
invalid-pose rates up to 100\,\% and every C3 episode ending in a stop
violation. A6 differs from A5 in exactly one respect, i.e. balanced
distillation inserted before PTQ, and it then retains C0 through C2 in full.
Thus, under the tested pathways, distilling before quantizing preserved
substantially more capability than quantizing the pruned actor directly. We
remind the readers that this is a placement observation on fixed pathways, and
not a factorial proof about operation ordering.

The most direct comparison in this study is A3 against A6, since the two share
every byte of their FP32 source configuration, and no training separates them.
While A3 passes all five curricula, its PTQ conversion A6 fails C3 and C4,
completing on C3 three of eight episodes where the FP32 source completed all
eight. 
The failure observed here is not an unsafe driving behavior, since A6 records
zero stop violations. It stops correctly and then never issues a driving
command again, remaining at the line until the episode times out. We note that
the zero-velocity commands are the emitted actions of the network itself,
reproduced bit for bit across repeats, and not a runtime fault.

\begin{figure}[htbp]
\centering
\includegraphics[width=\linewidth]{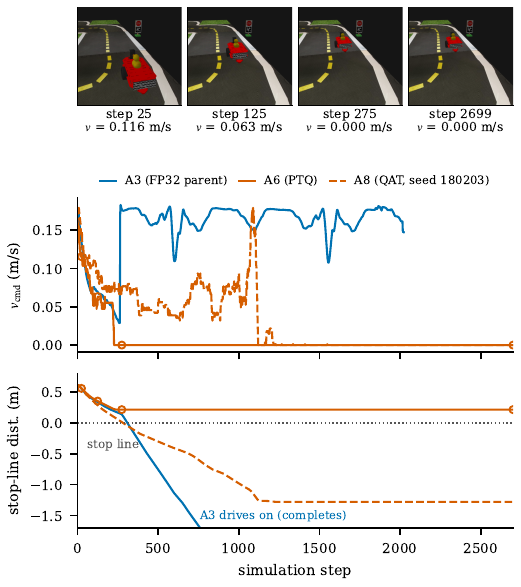}
\caption{The two quantized branches fail the stop curriculum from opposite
sides. Top: camera frames from the A6 episode itself. Below: commanded velocity and stop-line distance per step from the telemetry. }
\label{fig:phenotypes}
\end{figure}

The quantization-aware training approach appears to not bring back what post-training quantization has lost. This is because A8, obtained from
the same A3 configuration with quantization-aware training and teacher
guidance, also fails C3 and C4, and fails in the opposite direction. While A6
remains at the stop line, A8 drives through it, violating the stop in four of
eight C3 episodes and seven of eight C4 episodes. Furthermore, A8 never
retrains the quantized graph of A6, so that this is a comparison of two
quantization routes from one source configuration. Both quantized branches fail
the two curricula that require the policy to interrupt and resume its own
driving. They fail from opposite sides of that requirement. The failure
rows of Table~\ref{tab:failures} show the contrast directly, since A6 emits
$v_{\mathrm{cmd}}=0.000$ while parked 0.2\,m before the line, whereas A8 emits
$v_{\mathrm{cmd}}=0.077$ while crossing it. In Fig.~\ref{fig:phenotypes} we
show
the same contrast against their common source configuration.

\begin{table*}[htbp]
\caption{State and action at the first
failure. A dash in the stop-line column means no stop line
involved. Stop violations highlighted.}
\label{tab:failures}
\centering
\small
\setlength{\tabcolsep}{10pt}
\begin{tabular}{llclccc}
\hline
\textbf{Candidate} & \textbf{Task} & \textbf{Step/Horizon} & \textbf{Failure} & $\mathbf{v_{\mathrm{act}}}$ \textbf{(m/s)}
 & \textbf{Stop-line dist. (m)} & $\mathbf{v_{\mathrm{cmd}}}$ \textbf{(m/s)} \\
\hline
A1 & C0 & 367/368 & invalid pose & 0.135 & --- & 0.195 \\
A1 & C1 & 1312/1313 & invalid pose & 0.135 & --- & 0.195 \\
A1 & C2 & 248/249 & invalid pose & 0.136 & --- & 0.192 \\
\rowcolor{stoprow} A1 & C3 & 149/398 & stop violation & 0.148 & $-$0.001 & 0.219 \\
A1 & C4 & 4199/4200 & timeout & 0.000 & 0.451 & 0.000 \\
A2 & C0 & 126/127 & invalid pose & 0.242 & --- & 0.342 \\
A2 & C1 & 179/180 & lane failure & 0.228 & --- & 0.327 \\
A2 & C2 & 315/316 & lane failure & 0.139 & --- & 0.204 \\
A5 & C0 & 360/361 & invalid pose & 0.132 & --- & 0.196 \\
A5 & C1 & 1314/1315 & invalid pose & 0.134 & --- & 0.188 \\
A5 & C2 & 430/431 & invalid pose & 0.157 & --- & 0.225 \\
\rowcolor{stoprow} A5 & C3 & 150/932 & stop violation & 0.145 & $-$0.000 & 0.220 \\
A5 & C4 & 3227/3228 & lane failure & 0.022 & $-$0.896 & 0.032 \\
A6 & C3 & 2699/2700 & timeout (freeze) & 0.000 & 0.217 & 0.000 \\
A6 & C4 & 4199/4200 & timeout (freeze) & 0.000 & 0.187 & 0.000 \\
A7 & C0 & 128/129 & invalid pose & 0.244 & --- & 0.343 \\
A7 & C1 & 191/192 & invalid pose & 0.237 & --- & 0.357 \\
A7 & C2 & 309/310 & lane failure & 0.135 & --- & 0.203 \\
\rowcolor{stoprow} A8 & C3 & 284/2700 & stop violation & 0.051 & 0.001 & 0.077 \\
\rowcolor{stoprow} A8 & C4 & 1030/4200 & stop violation & 0.053 & $-$0.009 & 0.077 \\
\hline
\end{tabular}
\end{table*}

\subsection{Task Decision, Action Fidelity, and Cost}
\begin{figure*}[!t]
\centering
\includegraphics[width=\textwidth]{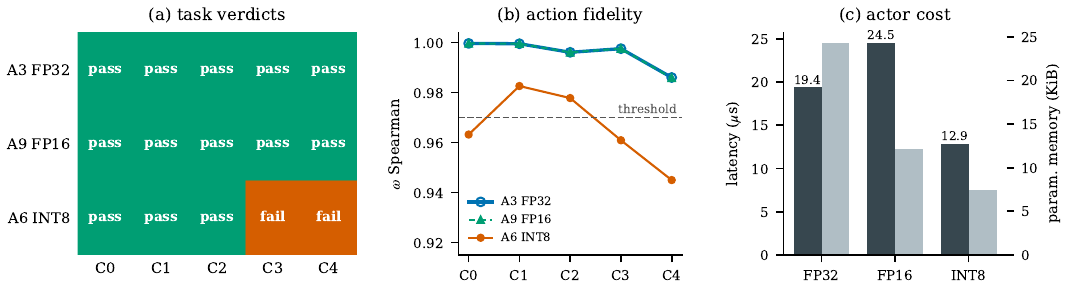}
\caption{(a) Task decisions for the FP32
source, its FP16 cast, and its INT8 PTQ conversion. (b) Spearman correlation
of the yaw-rate channel against the original actor. (c) Median
single-thread actor latency and logical parameter memory.}
\label{fig:precision}
\end{figure*}

Figure~\ref{fig:precision} compares the two precision routes that start from the
same A3 checkpoint. In terms of task decisions, the FP16 cast A9 passes all five
curricula, exactly as its FP32 source does, whereas the INT8 conversion A6 fails
C3 and C4. The action-fidelity panel shows the same separation, where the
Spearman correlation of the yaw-rate channel for A9 lies on top of the FP32
curve on every curriculum. Meanwhile the A6 correlation stays above the acceptance
threshold on C0 through C2 and drops below it on the stop curricula. In terms of
cost, the FP16 cast halves the parameter memory but adds 26\,\% to the median
actor latency over FP32 on this x86 backend, which is consistent with the
absence of a native half-precision compute path. The INT8 actor is the fastest
of the three at 12.9\,\textmu s against 19.4\,\textmu s for FP32, about
1.5$\times$ faster than its FP32 source, although its serialized file is larger
than the FP32 file, because the metadata of the traced quantized graph dominates
the weight savings at 6{,}210 parameters. On this backend, therefore, the
half-precision route preserves the driving capability at a latency penalty,
while the integer route buys speed at the cost of the two stop curricula.

\section{DISCUSSION AND LIMITATIONS}
\label{sec:discussion}

Our main results stem from three findings, which are a capability
loss with a known location (pruning), an improvement with an identified
operative factor (the rehearsal coverage), and a second and different failure
with an identified trigger (integer quantization of the narrow actor). None of
these findings is available from a report on the final configuration only. The stage-wise view also clarifies our position with respect to
accuracy-ordered compression, since our pipeline agrees with the
prune-before-quantize recommendation of
\cite{kim2026prunequantizeorder,shen2024order} but departs from the
distill-first prescription, i.e., using distillation to improve a lost capability and
not as an opening stage. 

Furthermore, both quantized branches fail exactly the curricula that require the policy to
interrupt and resume its own driving, and they fail from opposite sides. The
stop task makes the policy traverse a sharp decision boundary, from driving to
holding to driving again, under small input changes near the line. The behaviors of A6 and A8 are both consistent with small perturbations of
that boundary having large task consequences, whereas lane keeping tolerates
continuous small errors and survives. Our data locate this pattern but do not
establish its mechanism, so we make no claim about the internal representation
of the compressed networks.

Finally, the mismatch observed gives the fidelity-only
acceptance test both false negatives and false positives on the same small
model family. A8 would be accepted over A6 by fidelity, although A8 is the more
dangerous driver, while A4 would be rejected although it drives every
curriculum. We do not argue that fidelity metrics are useless, since they
remain cheap, differentiable, and useful in training. We argue instead that
closed-loop task acceptance cannot be inferred from them for a policy that must
act, and has to be measured.

\subsection{Implications for Deployment Practice}

Three practical consequences follow for researchers and engineers who aim to compress an automated
driving function before deployment. First, the release-gate evidence should be
produced at every intermediate stage and not only for the final configuration,
since a final-only report cannot separate a stage that removed a behavior from
a stage that failed to bring it back. Second, the rehearsal set used for any
recovery step is a safety-relevant configuration item, which deserves the same
coverage argument as the test set, since in our study it alone decided which
behaviors returned. Third, scenarios that require the vehicle to interrupt and
then resume its own driving should be treated as first-class acceptance
scenarios for compressed policies, since both of our integer routes failed
there while continuous lane keeping survived. All three consequences amount to
importing standard scenario-based assessment practice
\cite{riedmaier2020survey,ding2023survey} into the interior of the compression
pipeline.

\subsection{Limitations}
\label{sec:limitations}

We note a few limitations of our study. The acceptance checks in \eqref{eq:accept} are our operational definition of
the driving capability, so that a different check set could move individual
decision. We mitigated this by fixing every threshold before any result existed
and by never relaxing the deployment criterion later, but the definition
remains ad-hoc and not a standard. 
Furthermore, everything reported here also concerns one policy, one simulator, and one
compression toolchain, and the actor is small. Conclusions about a 6{,}210
parameter perceptron need not transfer to larger policies, where quantization
error could distribute differently. The integer-quantization procedure was also fixed by
design, so that the observed interaction is a property of the tested procedure
and not of INT8 as such, and the FP16 result carries its measured latency
caveat on the evaluated CPU. Finally, Gym-Duckietown remains a small-scale
platform, and the transfer of these findings to full-scale driving stacks is
untested, whether in higher-fidelity simulation or on the road. We aim to integrate higher fidelity simulation and on-track testing in our future work.

\section{CONCLUSION}
\label{sec:conclusion}

In this study, we process a belief-state visuomotor driving policy through structured pruning, knowledge distillation, integer quantization, and reduced floating-point inference. We evaluate every stage in closed loop and find that pruning is the stage at which the capability first broke, and the break covered every curriculum. Meanwhile, distillation brings part of the driving capability back exactly as far as its rehearsal data reached. Integer quantization of the improved actor loses the two hardest curricula again, through two opposite failure modes. Meanwhile, the same procedure on the unpruned width-256 actor preserves every curriculum, and so does a half-precision cast of the same checkpoint. We emphasize that action-level similarity predicted none of this in either direction. Our main conclusion is therefore not that any one technique is unsafe, but that a compression pipeline for a control policy is a sequence of behavioral transitions, whose surviving capability can only be established by making every intermediate configuration drive. Put together with the current streams of research on the design and the evaluation of automated driving systems, this endeavor would be highly beneficial to increase the implementability and to statistically guarantee the trustworthiness of compressed driving policies.
\bibliographystyle{IEEEtran}
\bibliography{references}

\end{document}